\documentclass[11pt]{article}

\usepackage[letterpaper,margin=1in]{geometry}
\usepackage{times}
\usepackage[round]{natbib}

\usepackage{amsmath,amsfonts,bm}

\def\eqref#1{equation~\ref{#1}}

\def\1{\bm{1}}

\DeclareMathAlphabet{\mathsfit}{\encodingdefault}{\sfdefault}{m}{sl}
\SetMathAlphabet{\mathsfit}{bold}{\encodingdefault}{\sfdefault}{bx}{n}

\usepackage{hyperref}
\usepackage{url}
\usepackage{graphicx}
\usepackage{booktabs}
\usepackage{colortbl,xcolor}
\definecolor{hlcol}{HTML}{EAF2FB}

\usepackage{fancyhdr}
\hypersetup{colorlinks=true,linkcolor=blue!50!black,citecolor=blue!50!black,urlcolor=blue!50!black}

\newcounter{algorithm}
\newcounter{algline}
\newcommand{\algrow}[1]{\stepcounter{algline}\thealgline & #1\\}
\newenvironment{algorithm}[1]{%
  \refstepcounter{algorithm}%
  \begin{center}\begin{minipage}{0.94\linewidth}%
  \smallskip\noindent\textbf{Algorithm~\thealgorithm\quad #1}\par\smallskip
}{%
  \smallskip\end{minipage}\end{center}%
}

\title{LAGO: Language Agents with Reusable Growing Options}

\author{
Yuran Wu$^{*,1}$ \quad
Mengkang Hu$^{*,2}$ \quad
Qionglin Qiu$^{3}$ \quad
Fanqing Meng$^{4}$ \\
Qiguang Chen$^{5}$ \quad
Yuhang Zhou$^{6}$ \quad
Qiudan Zhang$^{1}$ \quad
Xu Wang$^{\dagger,1}$ \\[0.8em]
\normalsize
$^{1}$Shenzhen University \quad
$^{2}$University of Hong Kong \quad
$^{3}$Hunan University \\
$^{4}$National University of Singapore \quad
$^{5}$Evolvent Research Team \quad
$^{6}$Fudan University
}
\date{}

\begin{document}

\maketitle

\renewcommand{\thefootnote}{\fnsymbol{footnote}}
\setcounter{footnote}{1}\footnotetext{Equal contribution.}
\setcounter{footnote}{2}\footnotetext{Corresponding author: \texttt{wangxu@szu.edu.cn}}
\renewcommand{\thefootnote}{\arabic{footnote}}
\setcounter{footnote}{0}

\begin{abstract}
Long-horizon tasks expose a weakness of monolithic tool-use loops: one agent must decompose, execute, and verify a task within a single trajectory. We introduce LAGO, which treats specialized language agents as reusable temporally extended options. A meta-agent retrieves options from a persistent library or spawns new ones, dispatches sub-tasks, verifies results, and admits successful novel options. The library grows across tasks without updating model weights. On 500 SWE-bench Verified tasks, frozen Qwen3.5-35B-A3B in both roles resolves 58.6\% under single-attempt evaluation with the official Docker evaluator. On a fixed 150-task subset, we re-run mini-SWE-agent, Terminus~2, and LAGO with the same policy, temperature, task IDs, and evaluator; their resolve rates are 49.3\%, 57.3\%, and 60.3\%, respectively. Cross-model experiments provide evidence consistent with a role for decomposition quality. A 50-task component ablation finds four fewer resolved tasks without decomposition, two fewer with shuffled reused content, and six fewer without the library. These results support reusable agent-options as a practical abstraction for task streams, while leaving component-level causality task- and model-dependent.
\end{abstract}

\section{Introduction}

Large language model agents have become a common interface for solving complex tasks with tools such as code execution, file editing, and web search. A dominant design pattern is the ReAct-style loop~\citep{yao2023react}, where a single agent repeatedly reasons, executes a tool action, observes the result, and continues until the task is completed. This design is simple and general, but it becomes brittle as task complexity increases. For realistic software engineering problems, an agent may need to understand a large repository, localize a bug across multiple files, implement a patch, run tests, and verify that no regression has been introduced. In such settings, a monolithic agent must simultaneously perform task decomposition, context retrieval, execution, verification, and recovery from errors. This often leads to context overflow, diluted attention over irrelevant observations, and error accumulation over long action trajectories.

A natural remedy is to hand-design a multi-agent pipeline. For example, different agents may be assigned specialized roles such as planner, coder, tester, reviewer, or researcher. Such systems can reduce the burden on any single agent by decomposing a complex task into smaller pieces. However, most existing multi-agent systems remain largely static: the roles, prompts, and interaction patterns are manually specified before execution. The system does not automatically acquire reusable capabilities from previous successful tasks. As a result, recurring sub-problems are repeatedly solved from scratch rather than being stored, verified, and reused.

We study how a language-agent system can acquire and reuse specialized capabilities across tasks. The system must decide when to reuse an existing agent, when to spawn a new one, how to verify and admit it into a persistent library, and how to limit redundancy.

We propose LAGO, a meta-agent framework that treats specialized language agents as temporally extended actions, or options. In hierarchical reinforcement learning, an option is a closed-loop policy that runs for multiple steps until a termination condition is met~\citep{sutton1999options,bacon2017optioncritic}. We adapt this abstraction to language agents. Each agent-option encapsulates a role specification, an intra-option policy implemented by an LLM-tool loop, a termination condition, and a verifier. The meta-agent operates over a growing action space consisting of primitive tools and reusable agent-options. During a task stream, it can retrieve existing options, spawn new ones when necessary, evaluate their outcomes, and maintain the library through admission and deduplication.

For complex agent tasks, performance may depend on the quality of the meta-agent's decomposition as well as the executor's capability. A strong executor can still fail when a task is split into overlapping, underspecified, or unverifiable sub-tasks. Better decompositions may help the same executor by isolating concerns, reducing context pressure, and enforcing verification.

We evaluate LAGO primarily on SWE-bench Verified with a single \emph{frozen} Qwen3.5-35B-A3B model in both roles, single-attempt sampling at temperature~$0$, and the official Docker evaluator. On the full 500-task split LAGO resolves 58.6\% of issues. Since the weights are frozen, this measures the delivered scaffold under a fixed backbone rather than separating framework and model effects. Published SWE-bench numbers are difficult to compare across papers---the same backbone spans 25\% to 62\% with different scaffolds and context budgets---so we re-run mini-SWE-agent and Terminus~2 end-to-end on the same frozen policy, 150-task subset, and temperature. We also hold the executor fixed while varying the meta-agent on SWE-bench Pro and BrowseComp, and measure library reuse in a separate rollout.

Our contributions are listed as follows:
\begin{itemize}
    \item We formulate specialized language agents as options, enabling a meta-agent to invoke temporally extended agent behaviors rather than only primitive tool calls.
    \item We introduce a growing agent-option library with spawn, admission, reuse, and deduplication mechanisms, allowing the effective action space to evolve over a task stream.
    \item We evaluate LAGO on SWE-bench Verified with a frozen Qwen3.5-35B-A3B as both meta-agent and executor under a single-attempt, temperature-$0$, official-Docker protocol: 58.6\% on the full 500-task split for absolute level, plus a controlled comparison in which two competing scaffolds are re-run end-to-end on the same frozen policy and the same 150-task subset. A supplementary cross-model analysis tests whether the observed differences are consistent with decomposition quality by holding the executor fixed and varying only the meta-agent.
\end{itemize}

\section{Related Work}

\textbf{Language agents and tool use.}
ReAct~\citep{yao2023react} interleaves reasoning and tool actions; related prompting and self-improvement methods include chain-of-thought~\citep{wei2022chain}, tree search~\citep{yao2023tot}, and self-reflection~\citep{shinn2023reflexion}. Tool-use agents range from direct tool calling~\citep{schick2023toolformer} to RL-based systems~\citep{singh2025artist,jin2025searchr1,wei2025swerl}; software-engineering agents apply them to repository-level repair~\citep{jimenez2024swebench,yang2024sweagent}. These systems often execute one long trajectory, leaving decomposition and verification inside the same context.

\textbf{Multi-agent systems and decomposition.}
Frameworks such as MetaGPT~\citep{hong2024metagpt}, AutoGen~\citep{wu2023autogen}, and CAMEL~\citep{li2023camel} coordinate fixed roles and prompts. Recent work learns orchestration policies~\citep{dang2025orchestration}; LAGO instead stores reusable agent-options and lets the action space change across tasks. This connects to self-evolving agents: Voyager~\citep{wang2023voyager} builds code skills, CODESKILL~\citep{li2026codeskill} evolves prompt-injected skills, and repository memory improves code localization~\citep{wang2025repomem}. Ratchet~\citep{ratchet2026} and Learning on the Job~\citep{learningonjob2026} study library hygiene and episode-derived rules. LAGO reuses complete specialized agents, each with its own specification, termination condition, and verifier.

\textbf{Options and hierarchical control.}
The options framework~\citep{sutton1999options} represents temporally extended closed-loop policies; Option-Critic~\citep{bacon2017optioncritic} and FeUdal Networks~\citep{vezhnevets2017feudal} learn or organize such policies. LAGO adapts this abstraction from motor control to language agents that execute multiple tool calls before returning control to a meta-agent.

\section{Method}
\label{sec:method}

\begin{figure}[t]
\centering
\includegraphics[width=0.98\textwidth]{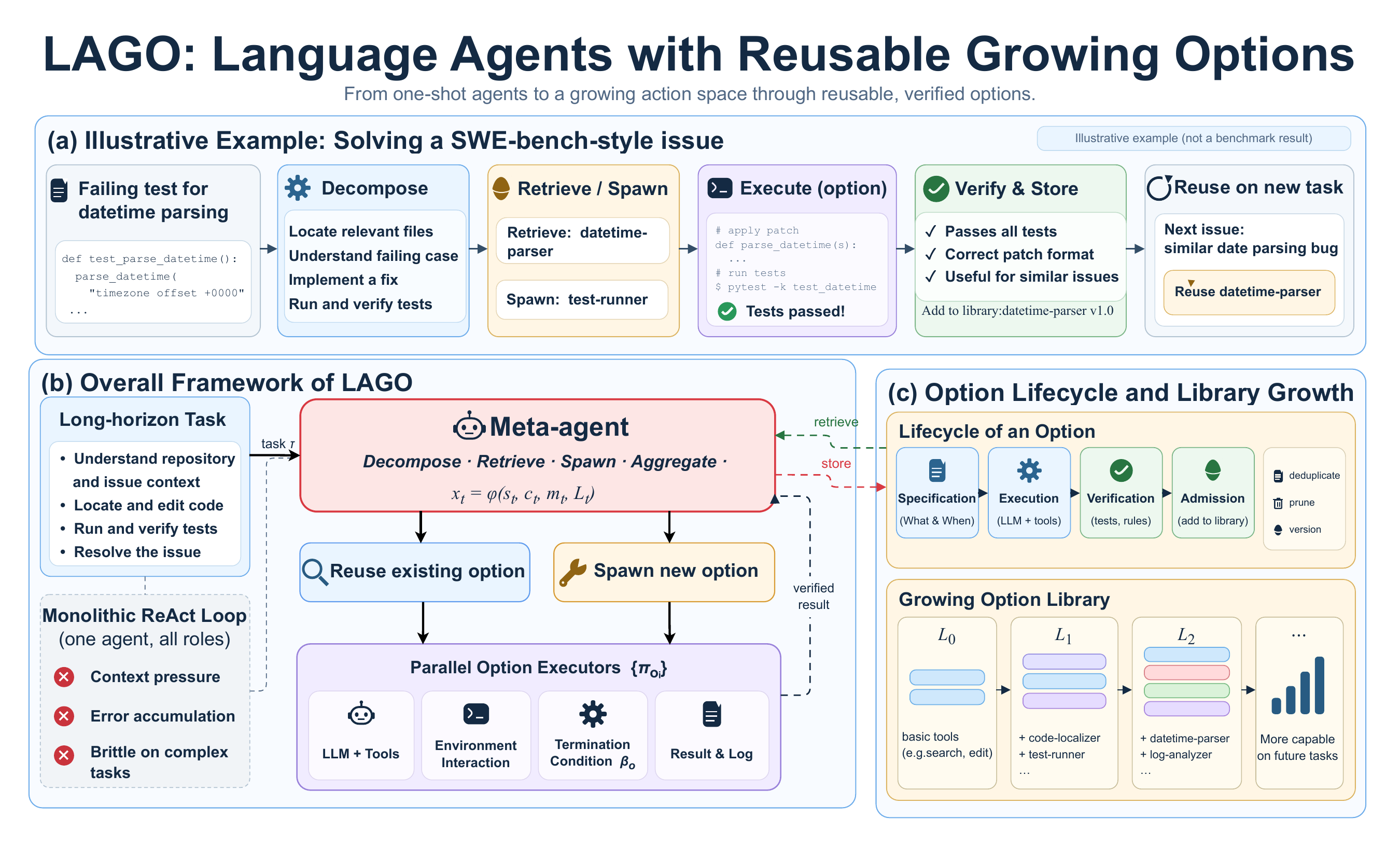}
\caption{LAGO turns task decomposition into a reusable action space. (a) An illustrative SWE-bench-style issue is decomposed, routed to reusable or newly spawned options, executed, verified, and reused on a later task. (b) The overall framework shows the meta-agent dispatching parallel options and receiving verified results. (c) The option lifecycle governs specification, execution, verification, admission, and deduplicated library growth from $L_0$ to $L_T$.}
\label{fig:architecture}
\end{figure}

\subsection{Overview}

LAGO uses a two-level agent architecture, illustrated in Figure~\ref{fig:architecture}. The meta-agent receives a user task, analyzes its complexity, decomposes it into sub-tasks, and decides whether each sub-task should be handled by an existing agent-option or by a newly spawned agent. Executor agents then solve the assigned sub-tasks through their own tool-use loops. The meta-agent aggregates the sub-results into a final patch or answer.

The design must decide when to spawn or reuse an option and how to admit, deduplicate, and prune library entries.

\subsection{Agents as Options}

We define an agent-option as a four-tuple:
\[
o = (\mathrm{spec}_o, \pi_o, \beta_o, v_o).
\]
Here, $\mathrm{spec}_o$ describes the role, goal, allowed tools, input-output contract, and execution limits of the sub-agent. $\pi_o$ is the intra-option policy implemented by an LLM and a tool-call loop. $\beta_o$ is the termination condition, which fires when the sub-task is completed, verified, failed, or its execution limit is reached. $v_o$ is a verifier that evaluates whether the option has produced a valid result.

When the meta-agent selects option $o$ at time $t$, the intra-option policy runs autonomously:
\[
a_{t+k} \sim \pi_o(\cdot \mid x_{t+k}), \quad \mathrm{until}\ \beta_o(x_{t+\tau}) = 1.
\]
The option then returns its result and verification signal to the meta-agent.

\subsection{Decision State and Action Space}

The meta-agent maintains a decision state:
\[
x_t = \phi(s_t, c_t, m_t, L_t),
\]
where $s_t$ is the task state, $c_t$ is the active context that fits within the prompt, $m_t$ is external persistent memory, and $L_t$ is the current agent-option library.

The action space at time $t$ is:
\[
A_t = A_{\mathrm{prim}} \cup O(L_t),
\]
where $A_{\mathrm{prim}}$ denotes primitive tool actions and $O(L_t)$ denotes reusable agent-options retrieved from the library. If no suitable option exists, LAGO falls back to ordinary tool-using behavior, so the library is not required for every task.

\subsection{Spawn, Admission, and Governance}

The meta-agent may execute a spawn action:
\[
\mathrm{spawn}(\mathrm{spec}) \rightarrow o_{\mathrm{new}}.
\]
A new option is created when the current library does not contain a suitable agent, when previous reuse attempts fail in a consistent way, or when the task appears to require a new skill. The newly spawned agent is evaluated before being admitted into the library.

We use an admission rule of the following form:
\[
\mathrm{admit}(o_{\mathrm{new}}) =
\mathbf{1}
[
v^{\mathrm{task}}(o_{\mathrm{new}}) \geq \theta_{\mathrm{task}}
\wedge
\mathrm{novelty}(o_{\mathrm{new}}, L_t) \geq \theta_{\mathrm{nov}}
].
\]
The task-level verifier checks whether the sub-task result is valid, whether the output satisfies the required format, and whether execution stays within its configured limit. The novelty term prevents near-duplicate agents from being repeatedly inserted into the library.

Library governance consists of three operations. First, deduplication merges agents with similar specifications and behavior, retaining the higher-quality validated version. Second, pruning removes agents that are rarely used or fail regression checks. Third, versioning records the prompt, tools, model version, and verifier version associated with each option.

\subsection{Objective over a Task Stream}

Over a stream of $T$ tasks, the system aims to maximize task success and quality while limiting library complexity:
\[
\max
\sum_{i=1}^{T}
\Big(
\mathrm{Success}(\tau_i)
+
\lambda \mathrm{Quality}(\tau_i)
\Big)
-
\rho |L_T|.
\]
The penalty on $|L_T|$ discourages unbounded library growth and encourages reuse over unnecessary spawning. Algorithm~\ref{alg:lago} in Appendix~\ref{app:algorithm} states the resulting inference loop over a task stream.

\subsection{Improving the Meta-Agent (Optional)}

In our main system the meta-agent is a frozen model and is not trained. Because decomposition quality is a central design concern, we additionally ask whether the meta-agent can be improved directly. To this end we fine-tune a Qwen3-30B meta-agent~\citep{yang2025qwen3} on successful decomposition traces and hold the executor fixed, so that the comparison tests a meta-agent change under a fixed executor. This is a supplementary analysis rather than a requirement of the framework: we report its effect in Section~\ref{sec:supp-ablation} and give training details in Appendix~\ref{app:sft-details}.

\section{Experiments}

\subsection{Experimental Setup}

We design the evaluation around three questions: (RQ1) under a strictly controlled protocol, does LAGO improve issue resolution over strong scaffolds re-run under the same protocol? (RQ2) are the observed differences consistent with decomposition quality across model families and task domains? (RQ3) is the growing agent library actively used and limited in net growth on the measured stream?

\label{sec:setup-protocol}
\textbf{Benchmarks and protocol.}
The main result is on SWE-bench Verified~\citep{jimenez2024swebench}, the 500-task human-validated split, where each task requires producing a patch for a real GitHub issue that passes the official test suites. Every SWE-bench Verified number we report is scored by the official Docker evaluator under the binary FAIL\_TO\_PASS/PASS\_TO\_PASS metric---no partial credit; the RQ2 benchmarks below use their own metrics. The controlled scaffold comparison (Table~\ref{tab:main-comparison-qwen}) runs on a fixed \textbf{150-task subset}, drawn once by repo-proportional stratified sampling (seed~42) and held identical across every system; re-running two external scaffolds end-to-end on the same local deployment is what makes the comparison controlled, and we use this fixed subset for the controlled comparison. At $n{=}150$ a resolve-rate difference carries roughly $\pm 8$ percentage points of binomial uncertainty, so we read only sizable gaps as meaningful. For absolute level we additionally evaluate LAGO on the \emph{full} 500 tasks, where it resolves 58.6\%; that figure shares the frozen-policy, single-attempt, official-evaluator protocol but is \emph{not} a row in the controlled table, because the external scaffolds were re-run only on the 150-task subset. For RQ2 we add two benchmarks of a different character: a 50-task subset of SWE-bench Pro~\citep{deng2025swebenchpro}, a longer-horizon repository-level extension of SWE-bench, and BrowseComp~\citep{wei2025browsecomp}, 50 English and 50 Chinese open-domain questions requiring multi-hop retrieval. Numbers on these two are \emph{not} comparable to the main result and are reported only to test cross-domain transfer.

\textbf{Systems.}
Our main system runs LAGO with a single \emph{frozen} Qwen3.5-35B-A3B~\citep{qwen2026qwen35} in both the meta-agent and executor roles; the weights are never updated. For the controlled comparison (Table~\ref{tab:main-comparison-qwen}) we re-run two training-free scaffolds end-to-end on this \emph{same} frozen policy rather than importing their published numbers: mini-SWE-agent, a minimal ReAct loop, and Terminus~2, a stronger terminal-agent scaffold (arXiv:2606.12882). Because the same weights and the same sampling temperature ($0$) drive every system, several sources of variation are shared; we do \emph{not} equalize the context window or the inference budget across scaffolds, and compare the delivered scaffolds under the shared task protocol. Methods that trained the policy with reinforcement learning are outside our (weight-frozen) scope and are excluded. For the RQ2 analysis we use a second family: a single-agent \textbf{Vanilla} baseline (one DeepSeek-V3.2~\citep{deepseekai2025v32} agent, no decomposition or library) versus LAGO variants that cross a meta-agent (DeepSeek-V3.2, Qwen3-30B base~\citep{yang2025qwen3}, or Qwen3-30B fine-tuned on successful decomposition traces, SFT) with an executor (DeepSeek-V3.2 or Qwen3-30B). Fixing the executor to DeepSeek-V3.2 and varying only the meta-agent tests whether the observed difference is consistent with decomposition quality. DeepSeek configurations used the provider's \texttt{deepseek-reasoner} endpoint (2026-02-17 to 2026-03-23), which its changelog resolves to DeepSeek-V3.2 in thinking mode; we report the underlying version.

\subsection{RQ1: LAGO Improves Resolution Under a Controlled Protocol}
\label{sec:main-comparison}

Table~\ref{tab:main-comparison-qwen} reports our controlled scaffold comparison on SWE-bench Verified. Published SWE-bench numbers are not comparable across papers: the same backbone spans 25\% to 62\% as the surrounding scaffold and context budget change. Rather than assemble a leaderboard from heterogeneous figures, we re-run each competing scaffold ourselves under the protocol of Section~\ref{sec:setup-protocol}, sharing weights, temperature, attempt count, task IDs, and evaluator. This is a comparison of delivered scaffolds under shared conditions; because context windows and inference budgets are not matched, it does not identify the effect of an isolated component. The two comparators span a strength progression: a minimal ReAct loop (mini-SWE-agent) and a stronger terminal-agent scaffold (Terminus~2), both of which report 59.8--61.6\% on this backbone at their native 200k window; our re-runs place them on LAGO's frozen policy and context budget instead, with the native-context figures retained for provenance in Appendix~\ref{app:reference-methods}.

\begin{table}[tb]
\centering
\small
\setlength{\tabcolsep}{5pt}
\begin{tabular}{l l r r}
\toprule
Method & Mechanism & Resolve (\%) $\uparrow$ & $\Delta$ vs. mini-SWE $\uparrow$ \\
\midrule
\multicolumn{4}{l}{\emph{Frozen Qwen3.5-35B-A3B; 150-task subset; temperature $0$; single attempt; official Docker}} \\
\midrule
mini-SWE-agent                            & single-agent ReAct loop   & 49.3$^{\mathrm{R}}$ & -- \\
Terminus~2                                & terminal-agent scaffold  & 57.3$^{\mathrm{R}}$ & +8.0 \\
\midrule
\rowcolor{hlcol}
\textbf{LAGO (ours)}                       & \textbf{hierarchical meta-agent with growing options} & \textbf{60.3$^{\mathrm{R}}$} & \textbf{+11.0} \\
\bottomrule
\end{tabular}
\caption{Controlled scaffold comparison on a \textbf{150-task subset} of SWE-bench Verified (repo-stratified, seed~42; identical task ids in every row). All rows use the same frozen Qwen3.5-35B-A3B at temperature~$0$, one attempt per task, and the official Docker evaluator. $\Delta$ is the descriptive percentage-point difference from mini-SWE-agent in this run. Each row is one end-to-end rerun ($^{\mathrm{R}}$) under the shared task protocol; Terminus~2 is run at temperature~$0$ for comparability.}
\label{tab:main-comparison-qwen}
\end{table}

\subsection{Component Ablation: Decomposition and Library Effects}
\label{sec:component-ablation}

To separate the contributions of decomposition and library reuse, we run three component ablations on a fixed 50-task split. The task IDs are stored in \texttt{eval/subset50\_abl\_ids.txt} and were obtained from \texttt{eval/subset150\_ids.txt} by repository-proportional stratification. There is no explicit runtime random seed: although \texttt{random.Random(42)} was used in the extraction script, the completed-task quota and sorted per-repository selection determine the final list, so the realized subset is deterministic. The six already-completed tasks fill their repository quotas first, and the remaining IDs are selected from the sorted candidates. All configurations use the same frozen Qwen3.5-35B-A3B served by local vLLM, temperature~$0$, one attempt, and a 65,536-token context budget with workspace truncation. All rates use the same denominator of 50; an unfinished end-to-end rollout is counted as unresolved. The full system resolves 30 tasks (60.0\%), while disabling decomposition resolves 26 tasks (52.0\%). Shuffling the content supplied by reused options lowers the result to 28 tasks (56.0\%), and disabling the library lowers it to 24 tasks (48.0\%). The corresponding approximate 95\% binomial intervals are reported in Table~\ref{tab:component-ablation}.

\begin{table}[tb]
\centering
\small
\setlength{\tabcolsep}{5pt}
\begin{tabular}{l c c c}
\toprule
Configuration & Resolved & Resolve (\%) & $\Delta$ vs. full \\
\midrule
LAGO (full) & 30/50 & $60.0\pm13.6$ & -- \\
Disable decomposition & 26/50 & $52.0\pm13.8$ & $-4$ \\
Shuffle reused content & 28/50 & $56.0\pm13.8$ & $-2$ \\
Disable library & 24/50 & $48.0\pm13.8$ & $-6$ \\
\bottomrule
\end{tabular}
\caption{Component ablation on a fixed 50-task split. All rows use denominator 50; unfinished rollouts are unresolved in this end-to-end accounting. The $\pm$ values are approximate 95\% binomial intervals, and $\Delta$ is the absolute difference in resolved tasks.}
\label{tab:component-ablation}
\end{table}

The ablation gives a bounded interpretation of the mechanism. On this split, disabling decomposition removes four resolved tasks, disabling the library removes six resolved tasks, and shuffling reused content has a smaller two-task effect. Thus the result supports a role for library availability in this setting, but does not establish that decomposition alone is necessary for success. This is compatible with the cross-model SFT result below, which suggests that decomposition quality can matter under some model and task conditions.

\subsection{RQ2: Observed Differences Are Consistent with Decomposition Quality}
\label{sec:supp-ablation}

Table~\ref{tab:main-comparison-qwen} establishes the result on a single frozen backbone; we now test whether the observed ordering is specific to that backbone or to code by re-running the decomposition on a \emph{second} model family (DeepSeek-V3.2, Qwen3-30B) and on BrowseComp. Tables~\ref{tab:abl-swebench} and~\ref{tab:abl-browsecomp} vary the meta-agent and executor independently across code repair (SWE-bench Pro) and open-domain search (BrowseComp). These numbers use a different backbone and are \emph{not} comparable to the main table; their role is cross-backbone, cross-domain evidence about the observed ordering, not causal attribution or a leaderboard.

\begin{table}[tb]
\centering
\begin{minipage}[t]{0.49\linewidth}
\centering
\footnotesize
\setlength{\tabcolsep}{4pt}
\begin{tabular}{lcccc}
\toprule
Meta (Exec.) & Spawn & Reuse & Rate & /task \\
\midrule
DS (DS) & 24 & 89 & 78.8 & 1.78 \\
Qwen3 (DS) & 30 & 172 & 85.1 & 3.44 \\
Qwen3-SFT (DS) & 31 & 197 & 86.4 & 3.94 \\
\bottomrule
\end{tabular}
\caption{Agent-library reuse statistics over the 50-task stream, by meta-agent (executor in parentheses). Rate $=$ reuse\,/\,(spawn $+$ reuse); Reuse/task is the mean reuse count per task.}
\label{tab:reuse}
\end{minipage}
\hfill
\begin{minipage}[t]{0.49\linewidth}
\centering
\scriptsize
\setlength{\tabcolsep}{2pt}
\begin{tabular}{llc}
\toprule
Meta-Agent & Executor & Resolve (\%) $\uparrow$ \\
\midrule
\textit{Vanilla} & DeepSeek & 34.0 \\
\midrule
\rowcolor{hlcol}
DeepSeek-V3.2     & DeepSeek-V3.2 & \textbf{50.0} \\
Qwen3-30B SFT   & DeepSeek-V3.2 & 40.0 \\
Qwen3-30B Base  & DeepSeek-V3.2 & 34.0 \\
Qwen3-30B SFT   & Qwen3-30B & 6.0 \\
Qwen3-30B Base  & Qwen3-30B & 4.0 \\
\bottomrule
\end{tabular}
\caption{Ablation on the SWE-bench Pro 50-task subset. The scaffold is held fixed; only the meta-agent and executor vary. Best result in \colorbox{hlcol}{shaded}; rows below the rule are ordered by resolve rate.}
\label{tab:abl-swebench}
\end{minipage}\end{table}

\begin{table}[tb]
\centering
\begin{minipage}[t]{0.49\linewidth}
\centering
\footnotesize
\setlength{\tabcolsep}{3pt}
\begin{tabular}{llccc}
\toprule
& & \multicolumn{3}{c}{Accuracy (\%) $\uparrow$} \\
\cmidrule(lr){3-5}
Meta & Executor & EN & ZH & Avg. \\
\midrule
\textit{Vanilla} & DeepSeek & 22.0 & 44.0 & 33.0 \\
\midrule
\rowcolor{hlcol}
DeepSeek-V3.2     & DeepSeek-V3.2 & \textbf{32.0} & \textbf{60.0} & \textbf{46.0} \\
Qwen3-30B Base  & DeepSeek-V3.2 & 20.0 & 44.0 & 32.0 \\
Qwen3-30B SFT   & DeepSeek-V3.2 & 22.0 & 42.0 & 32.0 \\
Qwen3-30B SFT   & Qwen3-30B & 6.0 & 6.0 & 6.0 \\
Qwen3-30B Base  & Qwen3-30B & 2.0 & 8.0 & 5.0 \\
\bottomrule
\end{tabular}
\caption{Ablation on BrowseComp (50 EN + 50 ZH); Avg.\ is the mean of EN and ZH. Best result in \colorbox{hlcol}{shaded}.}
\label{tab:abl-browsecomp}
\end{minipage}\hfill
\begin{minipage}[t]{0.49\linewidth}
\centering
\footnotesize
\setlength{\tabcolsep}{3pt}
\begin{tabular}{lcc}
\toprule
 & SFT Solved & SFT Failed \\
\midrule
Base Solved & 15 & 2 \\
Base Failed & 5 & 28 \\
\bottomrule
\end{tabular}
\caption{Instance-level comparison between Qwen3-30B base and SFT meta-agents on the SWE-bench Pro 50-task split with the same DeepSeek executor.}
\label{tab:sft-analysis}
\end{minipage}
\end{table}

\textbf{Cross-model comparison.}
On SWE-bench Pro (Table~\ref{tab:abl-swebench}), LAGO lifts resolve rate from 34\% to 50\% over the single-agent baseline, a 16-point difference under this protocol. Holding the executor fixed at DeepSeek-V3.2 and changing only the meta-agent, the SFT Qwen3-30B meta-agent reaches 40\% versus 34\% for its base variant---a 6-point difference consistent with an effect of decomposition quality, although the delivered systems still differ in their prompts and control flow. When both roles use the smaller Qwen3-30B, absolute performance falls to 4--6\%, while SFT remains slightly above the base variant; decomposition quality does not replace executor capability.

\textbf{Search benchmark.}
On open-domain search (Table~\ref{tab:abl-browsecomp}), dual-DeepSeek-V3.2 LAGO raises average accuracy from 33\% to 46\% (+13 points), with gains in both English (22$\to$32) and Chinese (44$\to$60), while swapping in the Qwen3-30B meta-agent variants lowers accuracy. This second benchmark shows that the observed ordering is not confined to the code-repair subset; it does not by itself establish a domain-independent mechanism.

\textbf{Instance-level comparison.}
The per-instance contingency (Table~\ref{tab:sft-analysis} and Figure~\ref{fig:sft-analysis}, Appendix~\ref{app:additional}) shows the SFT meta-agent solving 20 tasks to the base's 17: five new successes against two regressions. The wins concentrate where the base meta-agent under-decomposes---skipping verification or fusing separate concerns into one broad sub-task---while the regressions come from over-specified scope that leads the executor to miss edge cases a broader decomposition would have caught.

\subsection{RQ3: The Agent Library Is Actively Used and Deduplicated}
\label{sec:library}

\begin{figure}[tb]
\centering
\begin{minipage}[t]{0.48\columnwidth}
\centering
\includegraphics[width=\linewidth]{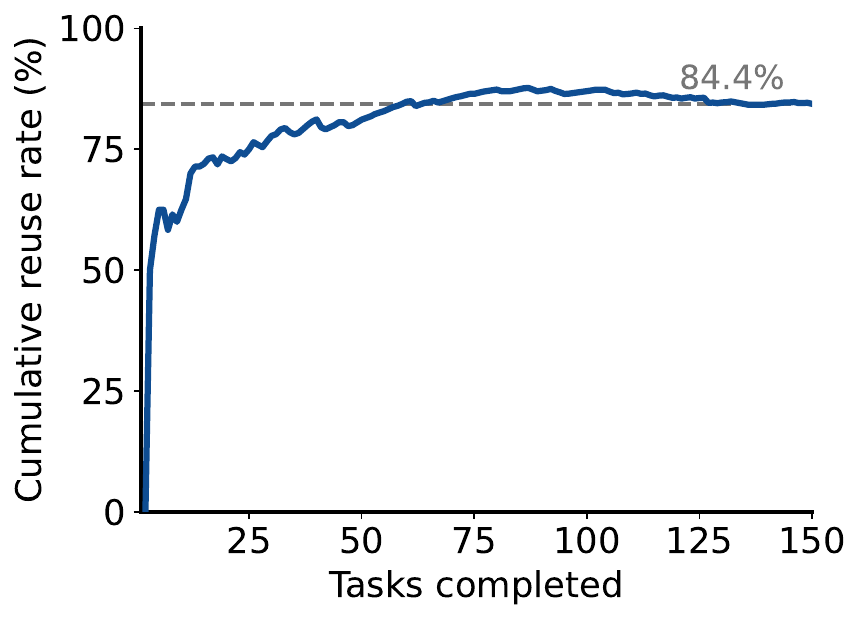}
\caption{Cumulative option reuse rate for a separate frozen Qwen3.5-35B-A3B, temperature-$0$, \texttt{max\_branch\_turns=150} rollout on the same 150 task IDs as the controlled comparison. It starts from the 123-option \texttt{agent\_library\_seed350} directory and is distinct from the separate 50-task reuse stream; final rate: 84.4\% (276 reuses / 51 spawns).}
\label{fig:reuse-cumulative}
\end{minipage}\hfill
\begin{minipage}[t]{0.48\columnwidth}
\centering
\includegraphics[width=\linewidth]{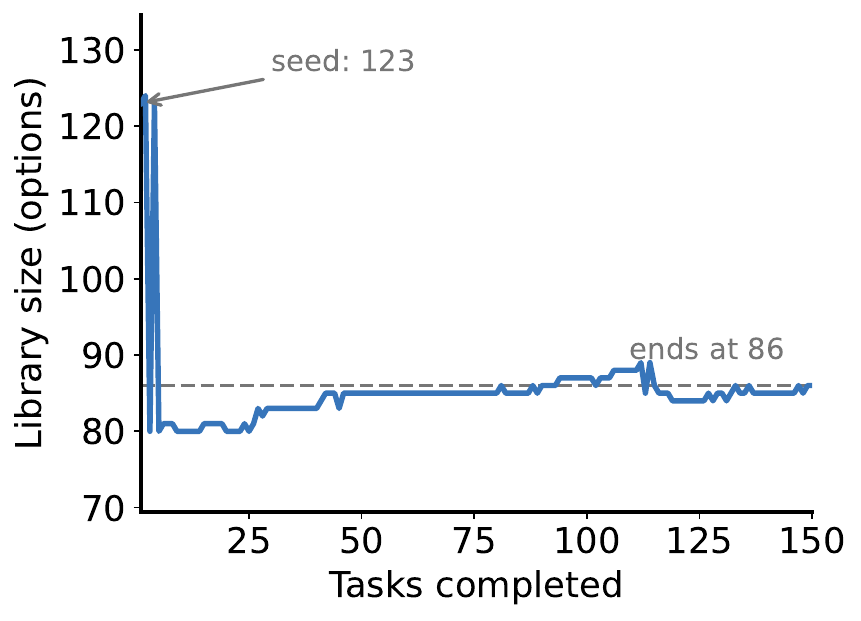}
\caption{Library size over the same rollout. The 123 options from \texttt{agent\_library\_seed350} are reduced to about 80 by initial deduplication and end at 86.}
\label{fig:library-size}
\end{minipage}
\end{figure}

Across all three systems, 78.8\%--86.4\% of executor options are retrieved from the library rather than freshly spawned (Table~\ref{tab:reuse}). In the separate 150-task rollout shown in Figures~\ref{fig:reuse-cumulative} and~\ref{fig:library-size}, cumulative reuse reaches 84.4\% (276 reuses and 51 spawns); the run is initialized from the 123-option \texttt{agent\_library\_seed350} directory, drops to about 80 after the initial deduplication, and ends at 86. This rollout uses the same 150 task IDs as the controlled scaffold subset but is a separate run. Reuse rate is the fraction of option executors sourced from the library, counted over executor children only (meta-level children and the root are excluded). It measures retrieval activity, not the causal value of retrieved content: retrieval uses a similarity function over option specifications, and separating transferred skill from generic prompt scaffolding requires a matched content ablation. The component ablation in Section~\ref{sec:component-ablation} provides partial evidence but does not fully isolate that effect.

Figure~\ref{fig:library-size} shows the corresponding library trajectory for the 150-task rollout: the seeded library starts at 123 options, drops to about 80 after the initial deduplication, and ends at 86. This is limited net growth after initialization for the measured rollout; it does not establish a long-run equilibrium across task distributions.

\section{Discussion}

Our cross-model results point to meta-agent decomposition quality as a plausible factor in some complex agent tasks. Poor decompositions fail in recognizable ways: overlapping sub-tasks that cause conflicting edits, missing verification that lets incorrect patches or unsupported answers pass, and scopes that are either too broad (raising context pressure) or too narrow (missing edge cases). SFT addresses some of these by producing smaller, more verifiable sub-tasks with clearer dependencies, but the regressions show that finer decomposition is not always better: a useful meta-agent must balance precision and coverage. The component ablation qualifies this interpretation: on its 50-task split, disabling decomposition removes four resolved tasks, so the marginal value of decomposition is task- and model-dependent.

The library is actively reused, and deduplication prevents the post-initialization stream from accumulating options unchecked. Reuse alone is not sufficient, however: the meta-agent must still recognize which option fits a given sub-task, so a strong library with a weak router can still fail.

Finally, the cross-model ordering separates two factors: dual DeepSeek is strongest overall, but the Qwen3 SFT meta-agent also improves over the Qwen3 base under the \emph{same} DeepSeek executor, so the ordering is consistent with a contribution from decomposition quality under a fixed executor. Improving agent systems may therefore involve decomposition, option selection, and library governance alongside executor capability.

\section{Limitations}
\label{sec:limitations}

\textbf{The controlled comparison is a 150-task subset.} At $n{=}150$ a resolve-rate difference carries roughly $\pm 8$ percentage points of binomial uncertainty, so only sizable gaps in Table~\ref{tab:main-comparison-qwen} are meaningful; the full-500 figure we report for absolute level was not accompanied by re-runs of the competing scaffolds, so the two cannot be merged into one ranking. The supplementary cross-model analysis and the component ablation use 50-task subsets and are correspondingly noisier; the component ablation also contains unfinished rollouts that are counted as unresolved under the fixed denominator.

\textbf{Reuse is measured, but not fully isolated.} We report how often options are retrieved (Section~\ref{sec:library}), which establishes that the library is active, \emph{not} that the retrieved content is what produces the gain. The component ablation (Section~\ref{sec:component-ablation}) partially addresses this: shuffling reused content reduces the result by two tasks and disabling the library reduces it by six on a 50-task split. These effects are noisy, and the shuffle does not match every property of the original option content; separating transferred skill from generic prompt scaffolding therefore remains incomplete. The same applies to spawning, deduplication, routing, and parallel versus sequential execution, none of which is individually isolated here.

Beyond these, library initialization must be controlled to avoid contamination between prior runs and evaluation tasks; we draw the seed library from a task set disjoint from the evaluated subset for this reason. Meta-agent fine-tuning is an optional extension with currently moderate gains, and more data is needed to characterize when it helps versus over-specifies. Finally, the verifier design is partly task-dependent, and more general verification is needed for broader deployment.

\section{Conclusion}

We introduced LAGO, a meta-agent framework that treats specialized language agents as temporally extended options: a meta-agent reuses existing options, spawns new ones, verifies outcomes, and maintains a growing library, turning task decomposition into a reusable component that requires no weight updates. On SWE-bench Verified a frozen Qwen3.5-35B-A3B in both roles resolves 58.6\% of the full 500 tasks, and a controlled comparison---two competing scaffolds re-run end-to-end on the same frozen policy, temperature, and 150-task subset---compares the delivered systems under shared conditions. A supplementary cross-model analysis on SWE-bench Pro and BrowseComp shows an ordering in which a stronger meta-agent performs better with the executor fixed, while the 50-task component ablation finds a four-task drop when decomposition is disabled and a six-task drop when the library is disabled. A separate library rollout shows high reuse with limited net growth after initialization.

The results suggest that decomposition is most useful when sub-problems are separable and recur across the task stream. Whether this pattern transfers to other growing-library agent systems remains an empirical question.

\subsection*{AI use statement}
During preparation of this manuscript, we used generative AI tools for code editing, figure scripting, language editing, and manuscript review. We did not use them to generate experimental results or citations. All AI-assisted outputs were reviewed by the authors, and code and figures were checked against the underlying data. The authors take responsibility for the final content, claims, and artifacts.

\bibliography{lago_iclr2027}
\bibliographystyle{plainnat}

\appendix

\section{Inference Loop}
\label{app:algorithm}

The loop below states the procedure described in Section~\ref{sec:method}; it is placed here to keep the main text on the mechanism and its evidence.

\begin{algorithm}{LAGO Inference over a Task Stream}
\label{alg:lago}
\setcounter{algline}{0}
\renewcommand{\arraystretch}{1.12}
\begin{tabular}{@{}r@{\quad}p{0.89\linewidth}@{}}
\toprule
\multicolumn{2}{@{}l@{}}{\textbf{Require:} task stream $\mathcal{D}$; initial option library $L_0$} \\
\multicolumn{2}{@{}l@{}}{\textbf{Ensure:} task results and updated library $L_T$} \\
\midrule
\algrow{Initialize $L \leftarrow L_0$.} 
\algrow{\textbf{for each} task $\tau \in \mathcal{D}$ \textbf{do}} 
\algrow{\quad Decompose $\tau$ into sub-tasks $\{u_j\}$.} 
\algrow{\quad\textbf{for each} sub-task $u_j$ \textbf{do}} 
\algrow{\quad\quad Retrieve candidate options from $L$.} 
\algrow{\quad\quad\textbf{if} a suitable option exists \textbf{then} reuse $o_j$;} 
\algrow{\quad\quad\textbf{else} spawn $o_{\mathrm{new}}$ from its specification.} 
\algrow{\quad\quad Execute the selected option and verify its result.} 
\algrow{\quad\quad\textbf{if} successful and novel \textbf{then} admit the option into $L$.} 
\algrow{\quad Aggregate verified sub-task results into the final answer or patch.} 
\algrow{\quad Periodically deduplicate, prune, and version $L$.} 
\algrow{\quad\textbf{end for}} 
\algrow{\textbf{end for}} 
\algrow{\textbf{return} results and $L_T$.} 
\bottomrule
\end{tabular}
\end{algorithm}

\section{Additional Experimental Results}
\label{app:additional}

This appendix collects supporting figures and tables for the supplementary cross-model analysis in Section~\ref{sec:supp-ablation}. They are referenced from the main text but placed here to keep the main-text evidence chain focused on the frozen Qwen3.5-35B-A3B comparison.

\begin{figure}[t]
\centering
\includegraphics[width=0.95\columnwidth]{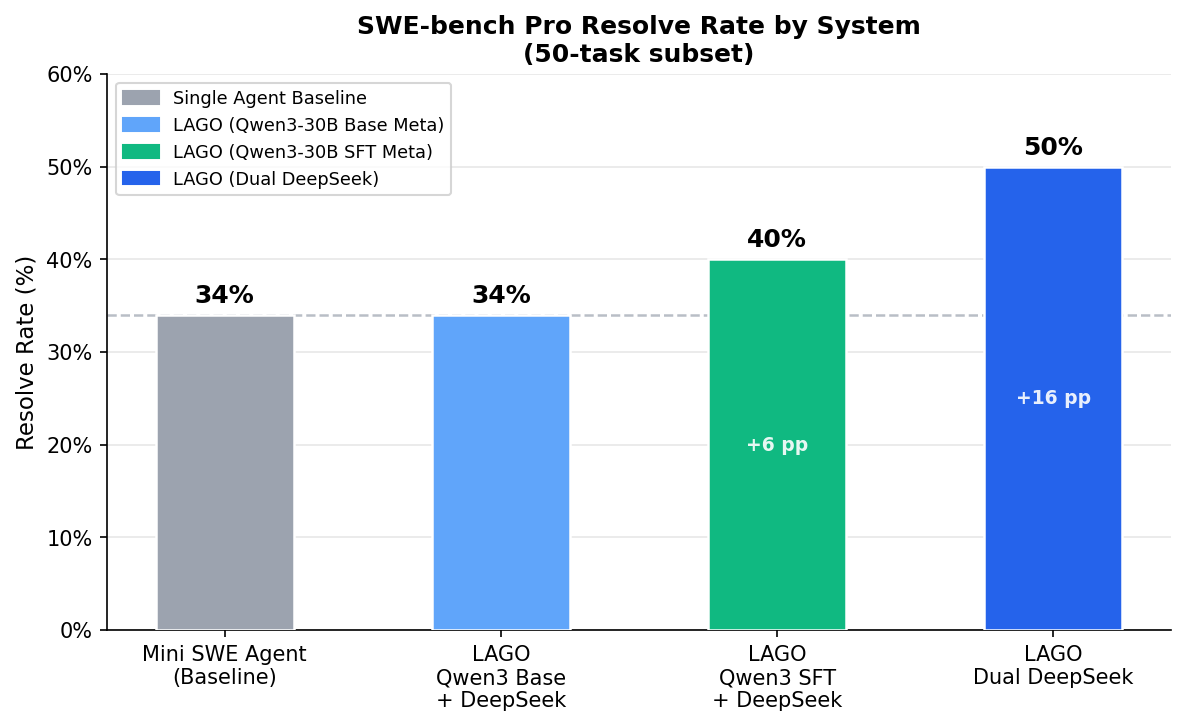}
\caption{SWE-bench Pro resolve rate by configuration (visualizing the numbers in Table~\ref{tab:abl-swebench}). LAGO with DeepSeek-V3.2 as both meta-agent and executor reaches 50\%, a 16-point gain over the single-agent baseline. With the executor fixed to DeepSeek-V3.2, an SFT Qwen3-30B meta-agent improves over the base variant by 6 points, while the base variant matches the baseline.}
\label{fig:main-results}
\end{figure}

\begin{figure}[t]
\centering
\includegraphics[width=0.95\columnwidth]{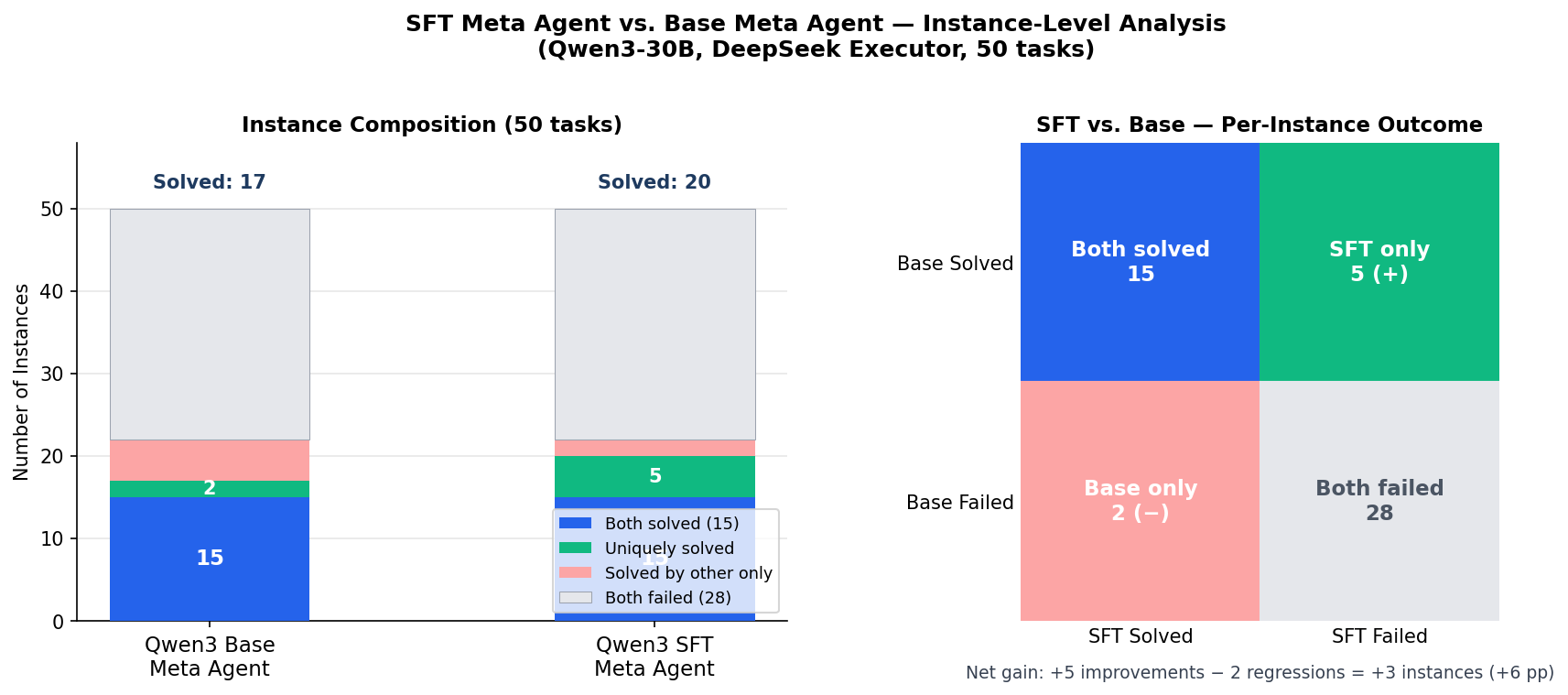}
\caption{Instance-level comparison of base and SFT Qwen3-30B meta-agents on the SWE-bench Pro 50-task split (DeepSeek executor). The two systems share 15 jointly-solved and 28 jointly-failed instances; SFT uniquely solves 5 and regresses on 2, for a net gain of 3 instances (17 to 20 solved).}
\label{fig:sft-analysis}
\end{figure}

\begin{figure}[t]
\centering
\includegraphics[width=0.95\columnwidth]{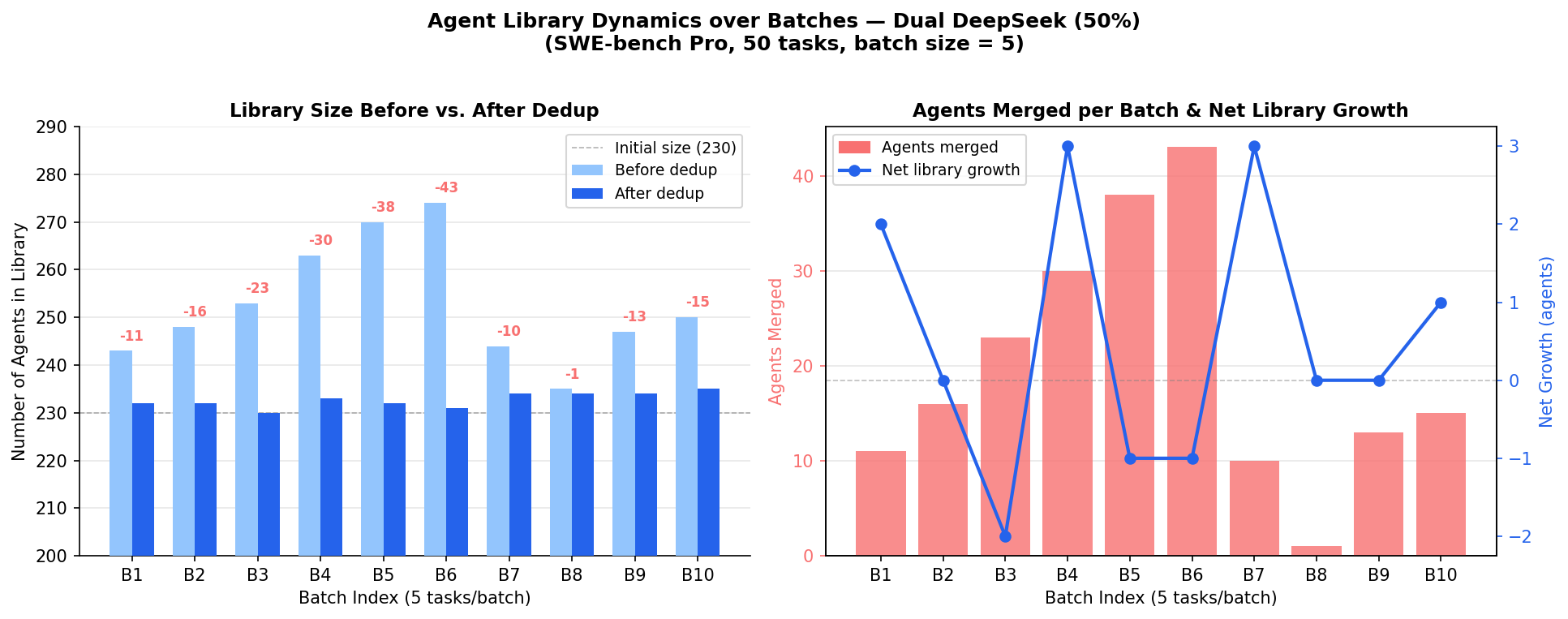}
\caption{Agent library dynamics for a separate 50-task Dual DeepSeek stream, distinct from the 150-task Qwen3.5-35B-A3B rollout in the main text. Left: library size before and after deduplication (red labels show agents merged). Right: agents merged and net growth per batch.}
\label{fig:library-growth}
\end{figure}

\begin{figure}[t]
\centering
\includegraphics[width=1\columnwidth]{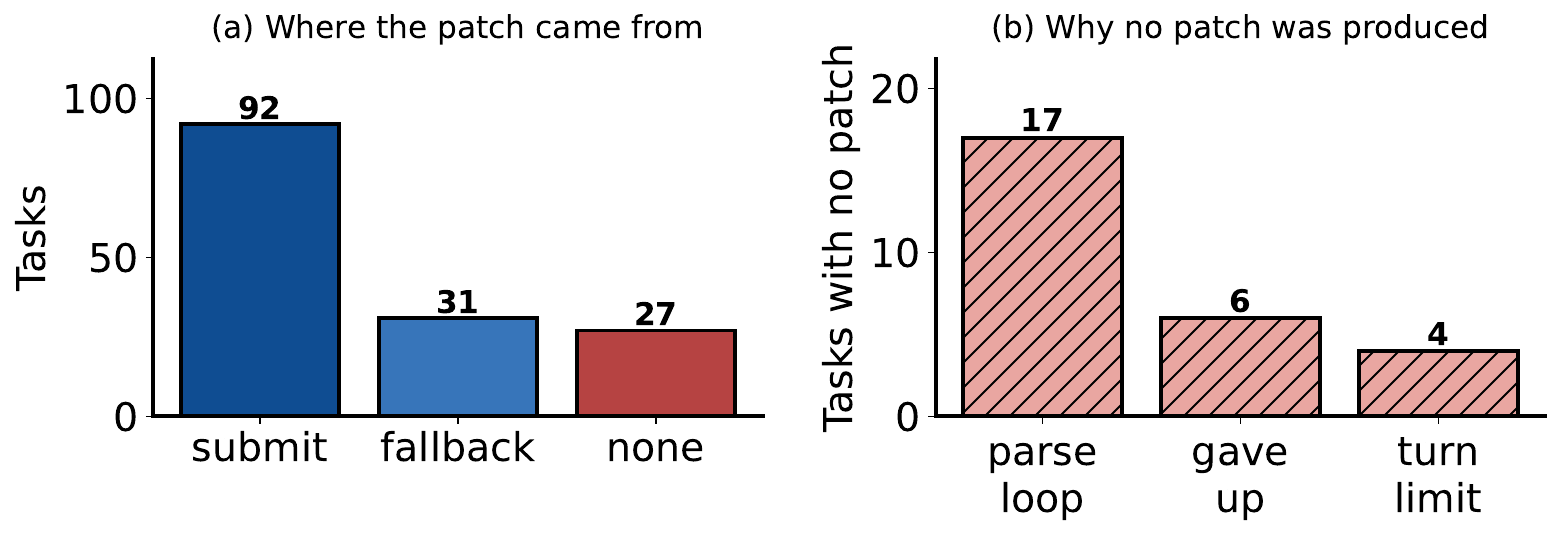}
\caption{Failure modes on the 150-task subset. (a)~Patch provenance: most tasks reach a \texttt{submit}, a further 31 are recovered by the unconditional final-diff fallback, and 27 end with no patch. (b)~Among those 27, the dominant causes are malformed-tool-call retries, abandonment after a sub-agent reaches its execution limit, and turn-limit exhaustion.}
\label{fig:failure-modes}
\end{figure}

\section{Reference Methods Under Non-Comparable Protocols}
\label{app:reference-methods}

Our main table (Table~\ref{tab:main-comparison-qwen}) is restricted to scaffolds we re-ran ourselves on a shared frozen policy, temperature, and task-id set. Several other recent training-free methods also use a Qwen3.5-35B-A3B policy, but each reports on its own task selection and grading setup, so their numbers are not interchangeable with ours and we do not rank them; Table~\ref{tab:reference-methods} lists them for provenance with the evaluated subset size made explicit. Two differences in particular tend to raise a reported number relative to our binary, task-id-matched metric: a hand-selected rather than stratified subset, and---for CODESKILL---a soft evaluator that grants partial credit where ours requires all held-out tests to pass.

\begin{table}[tb]
\centering
\small
\setlength{\tabcolsep}{4pt}
\begin{tabular}{lccp{5.4cm}}
\toprule
Method & Eval set & Orig.\ (\%) & Setup note \\
\midrule
Self-Harness                      & 100 tasks & 41.5      & self-modifying harness; mean of 2 runs; 67 held-in / 33 held-out \\
CODESKILL~\citep{li2026codeskill} & 150 tasks & 57.3--66.0 & frozen policy $+$ trained 4B skill manager; soft evaluator $V(\tau)\in[0,1]$ \\
CoACT                             & 200 tasks & 57.0--60.5 & mini-SWE-agent $+$ observation compression \\
TDAD                              & 25 tasks  & 24.0--32.0 & 4-bit quantized policy $+$ OpenCode; test-driven skill \\
\bottomrule
\end{tabular}
\caption{Training-free (weight-frozen) methods on SWE-bench Verified that use the same Qwen3.5-35B-A3B policy but evaluate only on a \emph{subset}, listed for provenance and not ranked against our re-run results (Self-Harness: arXiv:2606.09498, v3; CODESKILL: arXiv:2605.25430; CoACT: arXiv:2607.02911; TDAD: arXiv:2603.17973, whose 25-task stage uses a 4-bit quantized policy---its 100-task stage uses Qwen3-Coder-30B and is not this model). Ranges span each method's reported variants. RL-trained methods (e.g., LEGO-RL, CompactionRL) are outside our weight-frozen scope and omitted.}
\label{tab:reference-methods}
\end{table}

\paragraph{Native-context figures for the long-context scaffolds.}
The main table (Table~\ref{tab:main-comparison-qwen}) re-runs mini-SWE-agent and Terminus~2 on our frozen policy. For provenance, their \emph{native}-context results on this backbone (as reported by HarnessBridge, arXiv:2606.12882) are 59.8--61.6\% at a 200k effective window. Because deployment, context, and inference conditions differ, the gap between those native numbers and the re-run figures does not isolate the effect of context length; the native-context figures are retained only for provenance.

\section{Meta-Agent Fine-Tuning Details}
\label{app:sft-details}

For the supplementary meta-agent analysis in Section~\ref{sec:supp-ablation}, we fine-tune a Qwen3-30B meta-agent~\citep{yang2025qwen3} with supervised fine-tuning using 4-bit QLoRA~\citep{dettmers2023qlora}. The training traces are filtered by task success and decomposition quality, and each trace contains the task, the meta-agent's decomposition, sub-task specifications, and verification-oriented instructions.

The fine-tuning targets three behaviors. First, it encourages step completeness, especially the inclusion of verification and testing steps. Second, it improves chain refinement by producing smaller and clearer sub-tasks with explicit dependencies. Third, it improves task naming precision, producing agent specifications that are easier to retrieve and reuse. Fine-tuning can also introduce regressions when the generated sub-task scope becomes too narrow and misses edge cases, an effect we analyze at the instance level in Section~\ref{sec:supp-ablation}.

\end{document}